# GEOMETRIC FEATURE BASED FACE-SKETCH RECOGNITION

Sourav Pramanik
Computer Science and Engineering Department
National Institute of Science and Technology
Berhampur, India
srv.pramanik03327@gmail.com

Debotosh Bhattacharjee
Computer Science and Engineering Department
Jadavpur University
Kolkata, India
debotosh@ieee.org

*Abstract*— **This paper presents a novel facial sketch image or face-sketch recognition approach based on facial feature extraction. To recognize a face-sketch, we have concentrated on a set of geometric face features like eyes, nose, eyebrows, lips, etc and their length and width ratio because it is difficult to match photos and sketches because they belong to two different modalities. In this system, first the facial features/components from training images are extracted, then ratios of length, width, and area etc. are calculated and those are stored as feature vectors for individual images. After that the mean feature vectors are computed and subtracted from each feature vector for centering of the feature vectors. In the next phase, feature vector for the incoming probe face-sketch is also computed in similar fashion. Here, K-NN classifier is used to recognize probe face-sketch. It is experimentally verified that the proposed method is robust against faces are in a frontal pose, with normal lighting and neutral expression and have no occlusions. The experiment has been conducted with 80 male and female face images from different face databases. It has useful applications for both law enforcement and digital entertainment.**

## I. INTRODUCTION

The face identification problem is one of accepting or rejecting a person's claimed identity by searching from an existing face database to validate input. Face recognition has been an active research area over last 30 years but in recent years face sketch recognition has become an active research area for engineers or scientist because sketches are much different from photos in texture and shape and it is difficult to match photos and sketches as they are found in two different modalities [8]. Automatic retrieval of photos of suspects from the police mug shot database can help the police narrow down potential suspects quickly. However in most cases, the photo image of a suspect is not available. The best substitute is often a sketch drawing based on the recollection of an eyewitness. Therefore, automatically searching through a photo database using a sketch drawing becomes important. It can not only help police locate a group of potential suspects, but also help the witness and the artist interactively to modify the sketch during drawing based on similar photos retrieved [1], [2], [3], [4], [5], [6], [7].

It is well known that viewing a person's eyes, mouth is essential to grasping the information and emotions they convey and face components like eyes, nose, eyebrows, lips, etc together describe the overall shape of the face. Due to great difference between face photo and face-sketch it is not possible to compare directly, therefore in this system comparison has been done only between facial components/features of face photos and face-sketches. A lot of works on facial recognition and facial feature extraction have been reported in [9], [10], [11], [13], [14], [18], [15], [12]. However, due to the great difference between sketches and photos and the unknown psychological mechanism of sketch generation, face sketch recognition is much harder than normal face recognition based on photo images. It is difficult to match photos and sketches in two different modalities.

There was only limited research work on face sketch recognition because this problem is more difficult than photo-based face recognition and no large face sketch database is available for experimental study. Methods directly using traditional photo-based face recognition techniques such as the eigenface methods [1] and the elastic graph matching methods [2] were tested on two very small sketch data sets with only 7 and 13 sketches, respectively. In [3] [4], a face sketch synthesis and recognition system using eigentransformation was proposed. In [5] proposed a nonlinear face sketch synthesis and recognition method. It followed the similar framework as in [3] [4]. The drawback of this approach is that the local patches are synthesized independently at a fixed scale and face structures in large scale, especially the face shape, cannot be well learned. In [6], [7] proposed an approach using an embedded hidden Markov model and a selective ensemble strategy to synthesize sketches from photos. The transformation was also applied to the whole face images and the hair region was excluded. In [8], proposed a face sketch synthesis and recognition approach based on local face structures at different scale using a Markov Random Fields model. But the drawback of this approach is that it requires a training set containing photo-sketch pairs. In [15], [12] proposed methods for extraction of facial features. In [16], proposed an example-based face cartoon generation system. It was also limited to the line drawings and required the perfect match between photos and line drawings in shape. These systems relied on the extraction of face shape using face alignment algorithms such as Active Appearance Model (AAM) [17]. These line drawings are less expressive than the sketches with shading texture.

In this paper, propose a new approach to recognize face-sketch based on facial features extraction like eyes, eyebrows, nose, and lips. It requires a training set containing face photos and faces to be studied are in a frontal pose, with normal lighting and neutral expression, and have no occlusions. Here we extracted some common features and some unique features from face-sketch and face photo. To recognize face sketch we have considered eight facial features and associated parameters. Here we have used a geometric model shown in fig.3 to predict the positions where the facial components may appear and then extracts the actual regions of the facial components by applying the proper algorithms over the area around the predicted region. To predict the facial components region, first find the eye ball row and based on eye ball row we can easily predict the facial components region. Since the appearances of facial components are different from each other, it is not possible to extract all the facial components with one algorithm. Therefore, we designed separate algorithm for each facial component. Except facial features, we have considered one important parameter within the face, which is the length between upper-lip and the nostril. Because some people have small nose and some people have large nose. So the length between upper lip and nostril will be different. During face sketch recognition stage, first facial features are extracted from training photos and then find their length or width or area. After that represent each image features as a vector, if there are **n** images then there will be **n** vectors and for each vector, the mean of the vectors are computed and subtracted from each feature vector to get a set of vectors called zero-mean vector. Second, facial features are extracted from input face sketch images in similar manner that of face photo images and stored as a vector. This is also centered with respect to mean computed earlier. Third, a well known classifier, K-NN classifier has been used in a straightforward way to recognize input sketch. Here, we have considered K=5 and Euclidean distance measure. The experiment has been conducted on 80 male and female face images from different face database. Section II describes the overall system design, details of the experiments conducted along with results are given in section III, and section IV concludes the paper.

## II. FACE SKETCH RECOGNITION USING FACIAL FEATURE EXTRACTION

In this section, the overall system description for sketch face recognition based on facial feature extraction is given. We divided this section into two parts: (i) facial features extraction and analysis and (ii) recognition task.

### A. Facial Features Extraction

To describe a person, often we use his/her characteristic features like eye, eyebrow, lips, nose, hair, face cutting, etc as overall description. The human face description that our system accepts has been determined by a psychological study. The study reveals that seven facial components, namely Face Cutting, Right eye, Right eyebrow, Left eye, Left eyebrow, Nose, Lip are generally referred in describing a human face [15]. In [12], proposed a system FIDA which maps images in an existing database to a 14-dimensional descriptive feature space using softmax regression. In [15], proposed a system FASY for construction of human faces from textual description where they have considered seven facial components like Face Cutting, Right eye, Right eyebrow, Left eye, Left eyebrow, Nose and Lip for construction of human face.

In the present work, we have considered eight facial features which are shown in below.

- Face Cutting: Area of face.
- Left Eye: Length, Width.
- Right Eye: Length, Width.
- Left Eyebrow: Length.
- Right Eyebrow: Length.
- Nose: Length.
- Lip: Length, Width, Area.
- Distance between upper-lip and nostril

*1) Preprocessing:* Before extraction of the facial features some preprocessing tasks on training photos and test sketch face are needed. In the preprocessing step, photos are in color, first converted the RGB color to the gray level images and all the faces are cropped with the dimension of $150 \times 200$ pixels. Fig.1 shows (a) color image, (b) corresponding gray image, (c) corresponding cropped image. Fig.2 shows (1) a face sketch, (2) corresponding cropped face sketch.

*2) Extraction methodology:* In the present work, to recognize a face-sketch, extraction of facial component from images of human faces is required. The appearance of facial components are different from each other. It is not possible to extract all the facial components with one algorithm. Therefore, we have designed separate algorithm for each facial component. The first and the most important step in facial component detection is to track the position of the eyes. Thereafter, the symmetry property of the face with respect to the eyes is used for tracking rest of the components like eyebrows, lips, and nose. Here we have used a geometric model shown in fig.3 to predict the region of interest or the approximate positions where the facial components may appear and then the actual regions of the facial components are extracted by applying the proper algorithms over the area around the predicted regions. Here, we have considered two points (x1, y1) and (x2, y2) as the co-ordinates of the top left corner and bottom right corner of the predicted rectangular region for each facial component. All the calculations for the predicted regions in the geometrical model, used in this paper, are represented with respect to the width W and the length L of the face, where W is defined in terms of the number of columns and L is defined in terms of the number of the rows. The algorithms for extraction of the facial features are as follows:

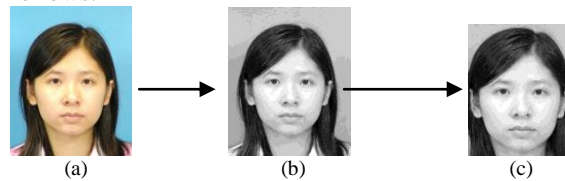

(a)　　　　　　(b)　　　　　　(c)

Fig.1 (a) Original color face photo (b) corresponding gray face photo (c) corresponding cropped face photo

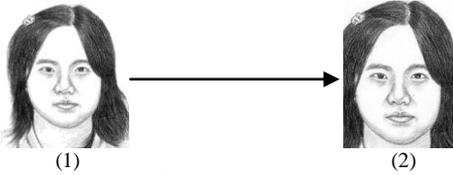

(1) (2)
Fig.2 (1) Original face-sketch, (2) Corresponding cropped image

*a) Face region extraction:* For extraction of face region, we have used mathematical morphology as a tool. The steps to extract the face region are as follows:-

    1. Convert faces into binary.
    2. Select structuring element (Here we have used Disk-shaped structuring element with radius 2).
    3. Apply Dilation operation.
    4. Apply Erosion operation.
    5. Apply Erosion operation.
    6. Apply Dilation operation.
    7. Replace all the background white pixels with black pixels.

*b) Right and left eye extraction:* For prediction of the eye-region, first extract the face region using above face region extraction method, after that we need to find the eye-ball row. Within the face region, we assume that minimum intensity values lay within the eye-ball. Because within the face intensity values (except eye-ball intensity) are must be greater than eye-ball intensity values. So if we add each row intensity values then the row where the eye-ball exists will have the minimum value among the rows. Therefore, based on minimum value row we can predict the eye region. The steps to extract the right eye are as follows:-

    1. Extract the face region.
    2. Find the row on which the eye ball exists (Called min_val_row).
    3. Now extract the portion of the image with the (x1, y1) and (x2,y2) values according to the geometric model. Where,

$$x1= min\_val\_row-c;$$
$$y1= a+1;$$
$$x2= x1+n-1;$$
$$y2= y1+m-1;$$

Where, the size of the sub-image is n × m.
The steps to extract the left eye are as follows:-

    1. Extract the face region.
    2. Find the row on which the eye ball exists (Called min_val_row).
    3. Now extract the portion of the image with the (x1, y1) and (x2, y2) values according to the geometric model. Where,

$$x1= min\_val\_row-c;$$
$$y1= W/2+b;$$
$$x2= x1+n-1;$$
$$y2= y1+m-1;$$

Where, the size of the sub-image is n × m.

*c) Right and left eyebrow extraction:* For extraction of eyebrow, we need to predict the eyebrow region by using the geometric model shown in fig 3. Here we need to first predict the eyeball row, then based on eyeball row we have predicted the eyebrow. The steps to extract the right eyebrow are as follows:-

    1. Extract the face region.
    2. Find the row on which the eye ball exists (called min_val_row).
    3. Now extract the portion of the image with the (x1, y1) and (x2, y2) values according to the geometric model. Where,

$$x1= min\_val\_row-(c*3);$$
$$y1= initial\_column + y;$$
$$x2= x1+n2-1;$$
$$y2= W/2-g;$$

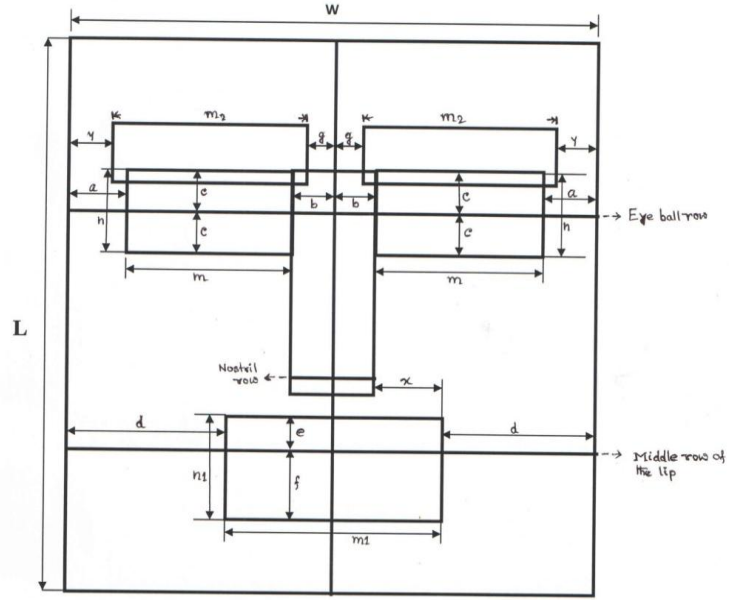

Fig.3 Geometric model used for this work

Where, W=150 column; L=200 row; n= 17 row; x= 10column; d= 50 column; y= 22 column; a= 30 column; e= 7 row; g= 7 row; b= 12 column; f= 17 row; c= 8 row; n1= 25 row; m=33 column; m1= 52 column;

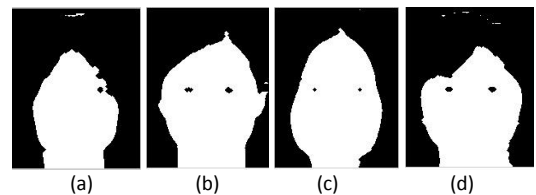

(a) (b) (c) (d)
Fig.4 (a), (b), (c), (d) extracted face region

The steps to extract the left eyebrow are as follows:-
    1. Extract the face region.
    2. Find the row on which the eye ball exists (called min_val_row).

3. Now extract the portion of the image with the (x1, y1) and (x2, y2) values according to the geometric model. Where,

$$x1 = min\_val\_row - (c*3);$$
$$y1 = W/2+g;$$
$$x2 = x1+n2-1;$$
$$y2 = y1+m2-1;$$

Where, the size of the sub-image is $n2 \times m2$.

*d) Lips extraction:* Like the previous methods, here also prediction of the lip region is done using the geometric model shown in Fig.3. The steps to extract the lip region are as follows:-

    1. Predict the row between two lips (called mid_lip_row).

    2. Now extract a portion of the photo with the (x1, y1) and (x2, y2) values according to the geometric model. Where,

$$x1 = mid\_lip\_row - e;$$
$$y1 = d+1;$$
$$x2 = x1+n1-1;$$
$$y2 = y1+m1-1;$$

Where, the size of the sub-image is $n1 \times m1$.

*e) Nose extraction:* For extraction of the nose, we need to predict the nose region by using the geometric model shown in Fig.3. Then we find the nostrils position for getting the lower boundary of the nose. Here have not considered the left or right boundary of the nose. Because in most of the faces the width of the nose is almost same. The steps to extract the nose are as follows:-

    1. Predict the nose region with (x1, y1) and (x2, y2) value according to the geometric model. Where,

$$x1 = R\text{-}EYE\_x1;$$
$$y1 = R\text{-}EYE\_y2;$$
$$x2 = x1+n2-1;$$
$$y2 = LIP\_y2 - 10;$$

Where, the length of the nose region is n2.

    2. Find the nostril position (called nostril_row**)**.

    3. Find the actual nose region with new (x1, y1) and (x2, y2) values. Where,

$$x1 = R\text{-}EYE\_x1;$$
$$y1 = R\text{-}EYE\_y2;$$
$$x2 = 1+(nostril\_row+2)-1;$$
$$y2 = 1+n2-1;$$

Where, R-EYE_x1 and R-EYE_y2 is the predicted right eye x1 and y2 co-ordinate and LIP_y2 is the predicted lip y2 co-ordinate.

*B. Facial component analysis*

In this phase, the analysis of the components is done. After extracting the actual region of the different facial components, we have analyzed each facial component to determine the length or width or area. After careful inspection of the images we found that the edge regions are darker from the other portion in the images. So if we add a large integer value (obviously between 0 and 255 because here we have used gray-level image) with each pixel intensity value of the image then the pixel intensities of the image other than edge region reaches nearly 255 or higher which can be considered as 255 also. Now, from this new image we can easily determine the edges and from this edge image we can find the length or width or area of the edge region. That is why, during the analysis process for the facial components we have normalized the predicted regions of the components by adding an integer 64 (one fourth of 255), where needed, with each intensity value of the components. After computation of the length or width or area, we take its ratio rather than actual length. For finding the length or width, we have found the edge image or binary image of each facial component. We have analyzed eight facial components and parameters for training face photo and input face-sketch separately. The algorithms for analysis of the facial features are as follows:

*1) Right and left eye length for face photo:* For finding the length of right or left eye, we need to first extract right or left eye. Using the above algorithm (discussed in section II.B) we have extracted right and left eye regions. The steps to find the length for right and left eye are as follows:

    i. Extract right and left eye region.
    ii. Find their binary image
    iii. Calculate their length from the binary image by finding the first black pixel and the last black pixel.
    iv. Compute their ratio.

*2) Right and left eyebrow length for face photo: :* For finding the length of right and left eyebrow, we need to first extract right and left eyebrow. The steps to find the length for right and left eyebrow are as follows:

    i. Extract right and left eyebrow region (called R_Ebrow and L_Ebrow).
    ii. Apply median filter on R_Ebrow and L_Eyebrow.
    iii. Normalize R_Ebrow and L_Eyebrow by adding an integer I (I=64) which is one fourth of 255.
    iv. Apply canny's edge detection algorithm to get edge image.
    v. Find their length from the edge image by finding the first white pixel and the last white pixel.
    vi. Take their ratio.

*3) Lip length for face photo and face sketch:* The steps to find the length and width for lip are as follows:

    i. Extract lip region (called LiP).
    ii. Apply median filter on LiP.
    iii. Normalize LiP by adding an integer I (I=64), which is one fourth of 255.
    iv. Apply canny's edge detection algorithm to get edge image.
    v. Find length and width from the edge image.
    vi. Take their ratio.

*4) Nose length for face photo:* The steps to find the length for nose are as follows:

    i. Extract nose region (called NOsE).
    ii. Find the nostril row from the extracted nose region (called nostril_row).
    iii. Extract actual nose region by adding 2 with nostril_row (called O_NOsE).

iv. Find the length.
v. Take the ratio.

*5) Right and left eye length for face-sketch:* The steps to find the length for right and left eye are as follows:
i. Extract right and left eye region.
ii. Find their binary image
iii. Calculate their length from the binary image by finding the first black pixels and the last black pixels.
iv. Take their ratio.

*6) Right and left eyebrow length for face-sketch:* The steps to find the length for right and left eyebrow are as follows:
i. Extract right and left eyebrow region (called R_Ebrow and L_Ebrow).
ii. Apply median filter on R_Ebrow and L_Eyebrow.
iii. Apply canny's edge detection algorithm to get edge image.
iv. Find their length from the edge image by finding the first white pixel and the last white pixel.
v. Take their ratio.

*7) Nose length for face-sketch:* The steps to find the length for nose are as follows:
i. Extract nose region (called NOsE).
ii. Find the nostril row from the extracted nose region (called nostril_row).
iii. Extract actual nose region (called O_NOsE).
iv. Find the length.
v. Take the ratio.

*8) Length between upper lip and nostril for face photo and face-sketch:* For finding the length between upper lip and nostril, we need to know the nostril position and the upper lip row. We have found nostril position when we have extracted actual nose region, and we can easily find the upper lip row from the edge image of lip. The steps to find the length between upper lips and nostril are as follows:
i. Find the nostril row from the actual nose region (called nostril_row).
ii. Find the upper lip row from the edge image of lips region (called u_lip_row).
iii. Find the length by taking the difference between u_lip_row and nostril_row, Length= u_lip_row – nostril_row.

## C. Recognition Task

After extraction and analysis of facial features from training images and incoming probe face-sketch, represent features as mean feature vectors for training images and mean feature vector for probe face-sketch then K-NN classifier is used to recognize probe face-sketch. The steps to recognition of a query sketch are as follows:

i) Obtain M training images $I_1, I_2, …, I_M$
ii) For each training image $I_i$ extract facial features or components $f_j$.
iii) For each training image $I_i$, find area or length or width ratio from the extracted features $f_j$ and represents it as a vector $\Gamma_i$.
iv) Find the mean $\Psi$ for every vector $\Gamma_i$ and subtract the mean from each feature vector $\Gamma_i$ to get a set of vector $\Phi_i$; $\Phi_i = \Psi - \Gamma_i$
v) Normalize incoming probe face sketch $\Omega$.
vi) Extract facial features or components $f_j$ from incoming test face sketch $\Omega$.
vii) Find the area or length or width ratio from the extracted features $f_j$ and represents it as a vector $\upsilon$.
viii) Find the mean $\rho$ for the vector $\upsilon$ and subtract the mean from the feature vector $\upsilon$ to get a new vector $\chi$; $\chi = \rho - \upsilon$
ix) Apply K-NN classifier to recognize incoming probe face sketch with K=5 and Euclidean Distance have used.
x) Display five face photo which are best matching with probe face-sketch.

## III. EXPERIMENTAL RESULTS

In this phase, we provide our experimental results. We have divided our experimental result into two parts. The first one shows the results for extraction of the facial components and the second one shows the recognize faces corresponding to the face-sketch.

### A. Experimental Result for extraction of the facial components

For extraction of the facial components, we have used a geometric method. We have tested this method on 80 male and female face images. The images are collected from CUHK training cropped photos face database. Fig. 5 shows some female and male face images collected from CUHK training cropped photos face database. Fig.7 to 11 shows the predicted region for some of the facial components for face photos.

Here we have also shown some face-sketches and their predicted region for various facial components. Fig.6 shows some male and female face-sketch collected from CUHK training cropped sketches database. Fig.12 to 16 shows the some of the predicted region for facial components for sketch faces. The performance measure for extraction of the facial components has been done by the equation 1.The success rate for extraction of facial components from face photo and face-sketch given in table-1.

Facial component extraction performance = $\frac{\text{Total successful extraction of a facial component}}{\text{Total no. of images} \times \text{Total no of components per image}} \times 100\%$ ………………………eq (1)

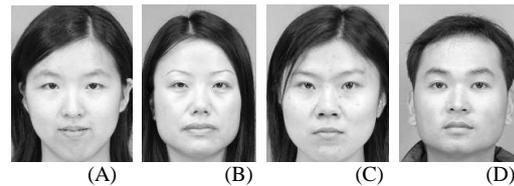

(A)    (B)    (C)    (D)
Fig.5 (A)-(D) Face photos from CUHK training cropped database

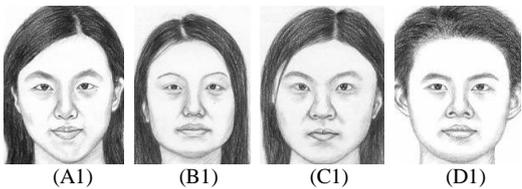
(A1)　　　(B1)　　　(C1)　　　(D1)
Fig.6 (A1)-(D1) face sketches from CUHK training cropped sketches database

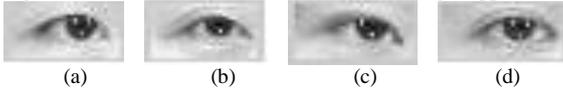
(a)　　　(b)　　　(c)　　　(d)
Fig.7 Extraction of right eyes from face photos

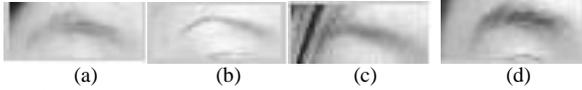
(a)　　　(b)　　　(c)　　　(d)
Fig.8 Extraction of right eyebrows from face photos

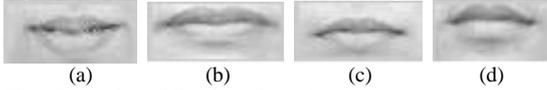
(a)　　　(b)　　　(c)　　　(d)
Fig.9 Extraction of lips from face photos

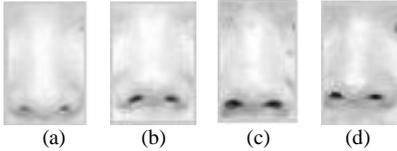
(a)　　　(b)　　　(c)　　　(d)
Fig.10 Extraction of predicted region of nose from face photos

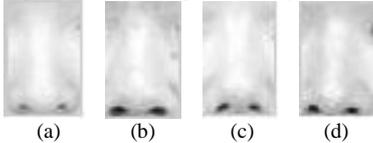
(a)　　　(b)　　　(c)　　　(d)
Fig.11 Extraction of actual nose region from predicted region

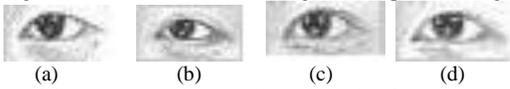
(a)　　　(b)　　　(c)　　　(d)
Fig.12 Extraction of left eyes from face-sketches

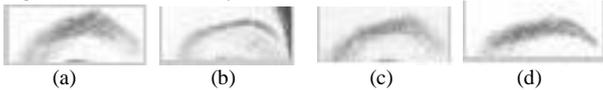
(a)　　　(b)　　　(c)　　　(d)
Fig.13 Extraction of left eyebrows from face-sketches

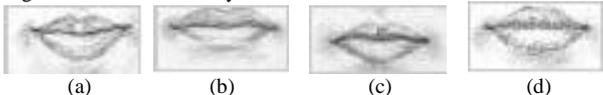
(a)　　　(b)　　　(c)　　　(d)
Fig.14 Extraction of lips from face-sketches

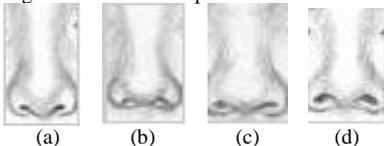
(a)　　　(b)　　　(c)　　　(d)
Fig.15 Extraction of predicted region of nose from face-sketches

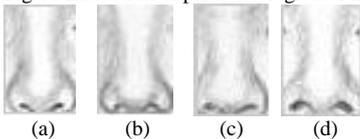
(a)　　　(b)　　　(c)　　　(d)
Fig.16 Extraction of actual nose region from predicted region

### B. Experimental result for Recognize a face sketch

In this section, we have given a snapshot to recognize a query face-sketch through face photo database. In our system, input is a face-sketch and output is five face photos which are best matching with input query face-sketch and rank them 1 to 5. In table-II, we compare our proposed method with Eigenface method and Sketch Transform Method [3]. The results in table-II clearly show the merit of our method over these two methods. The 1st match for Eigenface method is no more than 30% and 5th rank is no more than 60% and the 1st match for Sketch Transform Method is no more than 75% and 5th rank is no more than 90%. Our method greatly improves the 1st match to 80% and 5th match to 92.4%.

In Fig. 17, shows the query face-sketch and Fig. 18 shows the experimental result which shows five face photos (in order) which are best matching with the query face-sketch. In Fig.18, among the five face photos, original face photo corresponding to the query face sketch shows in the Figure 2 image (Rank 2).

## IV. CONCLUSION

In this paper, we have proposed a novel method to recognize a face sketch, based on extraction of facial components. This is different and difficult than face photo recognition because faces are much different from sketches in terms of color, texture, and projection details of 3D faces in 2D images. For extraction of facial components, we have used a geometric model which has been discussed in this paper. Here we have considered eight facial components and for extraction of each of those facial components we have designed distinct algorithm. After extraction of facial components, their length, width, and area are computed and then some specific ratios are computed to construct discriminating feature vectors. Finally, K-NN classifier has been employed to recognize probe face-sketch through face photos database. To validate this new approach, the approach was tested using CUHK face database.

TABLE I. SUCCESS RATES FOR FACIAL FEATURE EXTRACTION

| Features | Success rate of CUHK training cropped photos (%) | Success rate of CUHK training cropped sketches (%) |
|---|---|---|
| Right eye | 97 | 97 |
| Left eye | 97 | 97 |
| Right eye brow | 94 | 95 |
| Left eyebrow | 94 | 94 |
| Nose | 97 | 97 |
| Lips | 96 | 96 |

TABLE II. MATCHING PERCENTAGE OF THREE METHODS

| Rank | 1 | 2 | 3 | 4 | 5 |
|---|---|---|---|---|---|
| Eigenface | 26 | 33.8 | 47.1 | 52.4 | 55 |
| Sketch Transform Method | 71 | 78 | 81 | 84 | 88 |
| Proposed Method | 80 | 82.1 | 84 | 90.1 | 92.4 |

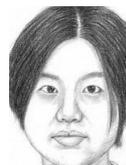
Fig. 17 query face-sketch

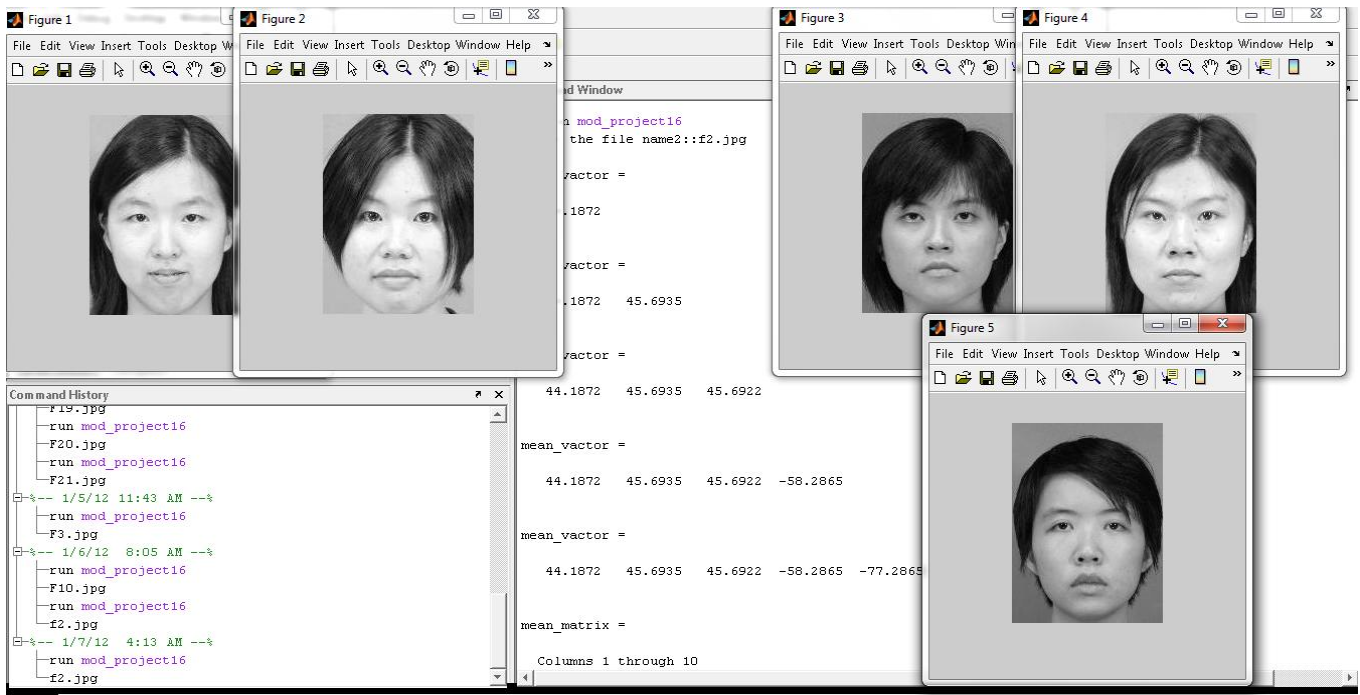

Fig. 18 snapshot to recognize query face-sketch

In our system, face-sketch is the input image and output is five face images which are best matching with input image. Curve-based feature extraction may be employed in future for finding better features and SVM classifier for better classification.

ACKNOWLEDGMENT

Authors are thankful to a major project entitled "Design and Development of Facial Thermogram Technology for Biometric Security System," funded by University Grants Commission (UGC),India and "DST-PURSE Programme" at Department of Computer Science and Engineering, Jadavpur University, India for providing necessary infrastructure to conduct experiments relation to this work.

REFERENCES

[1] R.G. Uhl and N.D.V. Lobo, "A Framework for Recognizing a Facial Image from a Police Sketch," Proc. IEEE Int'l Conf. Computer Vision and Pattern Recognition, 1996.

[2] W. Konen, "Comparing Facial Line Drawings with Gray-Level Images: A Case Study on Phantomas," Proc. Int'l Conf. Artificial Neural Networks, 1996.

[3] X. Tang and X. Wang, "Face Sketch Recognition," IEEE Trans. Circuits and Systems for Video Technology, vol. 14, no. 1, pp. 50-57, Jan. 2004..

[4] X. Tang and X. Wang, "Face Sketch Synthesis and Recognition," Proc. IEEE Int'l Conf. Computer Vision, 2003.

[5] Q. Liu, X. Tang, H. Jin, H. Lu, and S. Ma, "A Nonlinear Approach for Face Sketch Synthesis and Recognition," Proc. IEEE Int'l Conf. Computer Vision and Pattern Recognition, 2005.

[6] J. Zhong, X. Gao, and C. Tian, "Face Sketch Synthesis Using a E-Hmm and Selective Ensemble," Proc. IEEE Int'l Conf. Acoustics, Speech, and Signal Processing, 2007.

[7] X. Gao, J. Zhong, and C. Tian, "Sketch Synthesis Algorithm Based on E-Hmm and Selective Ensemble," IEEE Trans. Circuits and Systems for Video Technology, vol. 18, no. 4, pp. 487-496, Apr. 2008.

[8] X. Tang and X. Wang, "Face Photo-Sketch Synthesis and Recognition," IEEE Trans. Pattern analysis and Machine intelligence, vol. 31, no. 11,Nov. 2009.

[9] Wu H, Chen Q, Yachida M (1995) An application of fuzzy theory: face detection. In: Proceedings of IWAFGR'95, pp 314–319.

[10] Yang MH, Kriegman, DJ, Ahuja N (2002) Detecting faces in images: a survey. IEEE Trans Pattern Anal Mach Intell 24(1):34–58.

[11] Rowley HA, Baluja S, Kanade (1998) Neural network-based face detection. IEEE Trans Pattern Anal Mach Intell 20(1):23–38.

[12] Bhatia N, Kumar R, Menon S (2007) FIDA: face recognition using descriptive input semantics, December 14.

[13] Choi HC, Oh SY (2005) Face detection in static images using Bayesian discriminating feature and particle attractive genetic algorithm. In: Intelligent robots and systems (IROS2005), pp 1072–1077.

[14] Gu H, Su G, Du C (2003) Feature points extraction from faces. In: Image and vision computing (IVCNZ'03), pp 154–158.

[15] D. Bhattacharjee, S. Halder, M. Nasipuri, D.K. Basu, M. Kundu "Construction of human faces from textual description", Springer, Nov. 2009.

[16] H. Chen, Y. Xu, H. Shum, S. Zhu, and N. Zheng, "Example-Based Facial Sketch Generation with Non-Parametric Sampling," Proc. IEEE Int'l Conf. Computer Vision, 2001.

[17] T.F. Cootes, G.J. Edwards, and C.J. Taylor, "Active Appearance Model," Proc. European Conf. Computer Vision, 1998.

[18] Gilbert JM (1993) A real time face recognition system using custom VLSI hardware. Harvard Undergraduate Honors Thesis in Computer Science.